\documentclass[14pt]{extarticle}
\usepackage[letterpaper, margin=1in]{geometry}
  
\usepackage{graphicx}
\usepackage{amsmath}
\usepackage[hidelinks]{hyperref}
\usepackage{cleveref}
\usepackage{parskip}
\usepackage[square,numbers,sort&compress]{natbib}
\usepackage[charter]{mathdesign}
\usepackage{microtype}
\usepackage{xcolor}

\usepackage{csquotes}

\usepackage{authblk} % multiple authors

\usepackage{titlesec}
\titleformat{\section}{\normalsize\bfseries}{\thesection}{1em}{}
\titleformat{\subsection}{\normalsize\bfseries}{\thesubsection}{1em}{}
\titleformat{\subsubsection}{\normalsize\bfseries}{\thesubsubsection}{1em}{}
\usepackage[font=small]{caption}

\title{Active learning for regression in engineering populations: A risk-informed approach}

\author[1]{Daniel R. Clarkson}
\author[2]{Lawrence A. Bull}

\author[1]{Tina A. Dardeno}
\author[1]{Chandula T. Wickramarachchi}
\author[1]{Elizabeth J. Cross}
\author[1]{Timothy J. Rogers}
\author[1]{Keith Worden}
\author[1]{Nikolaos Dervilis}
\author[1]{Aidan J. Hughes}

\affil[1]{Dynamics Research Group, University of Sheffield, Mappin St, Sheffield, S1 3JD, UK}
\affil[2]{School of Mathematics and Statistics, University of Glasgow, Glasgow, G12 8SQ, UK}

\date{September 2024}

\begin{document}

\maketitle

\begin{abstract}
Regression is a fundamental prediction task common in data-centric engineering applications that involves learning mappings between continuous variables. In many engineering applications (e.g.\ structural health monitoring), feature-label pairs used to learn such mappings are of limited availability which hinders the effectiveness of traditional supervised machine learning approaches. The current paper proposes a methodology for overcoming the issue of data scarcity by combining active learning with hierarchical Bayesian modelling. Active learning is an approach for preferentially acquiring feature-label pairs in a resource-efficient manner.  In particular, the current work adopts a risk-informed approach that leverages contextual information associated with regression-based engineering decision-making tasks (e.g.\ inspection and maintenance). Hierarchical Bayesian modelling allow multiple related regression tasks to be learned over a population, capturing local and global effects. The information sharing facilitated by this modelling approach means that information acquired for one engineering system can improve predictive performance across the population. The proposed methodology is demonstrated using an experimental case study. Specifically, multiple regressions are performed over a population of machining tools, where the quantity of interest is the surface roughness of the workpieces. An inspection and maintenance decision process is defined using these regression tasks which is in turn used to construct the active-learning algorithm. The novel methodology proposed is benchmarked against an uninformed approach to label acquisition and independent modelling of the regression tasks. It is shown that the proposed approach has superior performance in terms of expected cost -- maintaining predictive performance while reducing the number of inspections required.

\end{abstract}

\vfill
\vspace{4em}

\section{Introduction}

Structural Health Monitoring (SHM) offers proactive solutions to ensure the safety and reliability of various items of infrastructure. SHM systems consist of data acquisition and processing systems to enable the detection of damage in monitored structures. The aim of SHM is to help inform decision-making, particularly for the operation and maintenance (O\&M) of high-value and safety-critical infrastructure. Improving decision-making in SHM has economic benefits, by reduction of unnecessary inspections and interventions, and safety benefits, by reducing the likelihood of failure events via informed interventions.

Statistical pattern recognition (SPR) is widely recognised as the primary tool for data-driven predictions in SHM systems \cite{farrarStructuralHealthMonitoring2012}. Regression models are a fundamental component of the SPR approach to decision-support technologies, enabling the prediction of continuous outcomes based on acquired data. For example, within the context of SHM, by associating the target variables of a regression model with salient health-states, inferences can be made regarding the condition of a structure of interest - although assumed as discrete in many SHM problems, damage progression is usually, in reality, a continuous variable.

A significant challenge in SHM is the scarcity of data. All data-driven models can suffer from bias and high uncertainty without sufficient training data, leading to unreliable predictions. Acquiring extensive labelled datasets capturing a structure's behavior across various health conditions is prohibitively costly and often unattainable for essential infrastructures. Inadequate data motivate sharing information between similar assets \cite{bullHierarchicalBayesianModeling2023a}. From this a new approach has emerged, Population-based structural health monitoring (PBSHM) \cite{bullFoundationsPopulationbasedSHM2021,gosligaFoundationsPopulationbasedSHM2021,gardnerFoundationsPopulationbasedSHM2021,tsialiamanisFoundationsPopulationbasedSHM2021}.

\subsection{Population-based Structural Health Monitoring}
PBSHM considers an entire population of structures. This approach assumes that structures that share common environmental conditions, load patterns, and/or aging effects and thus will share statistical commonalities. Considering structures as a population allows SHM users to share data between them. It allows a structure with rich historical data to lend its statistical strength to a data-poor structure. By capturing data from multiple structures, PBSHM aims to provide a more accurate and holistic health assessment and enables the identification of global patterns, trends and anomalies that would be difficult to observe at the individual level. This strategy can not only improve predictions for structures with very little data, but also makes the most out of comprehensive datasets that can be so costly to procure. PBSHM has a suite of tools that can be used to share information across a population, namely transfer learning and domain adaptation. For a thorough introduction to the population-based approach to SHM see \cite{bullFoundationsPopulationbasedSHM2021,gosligaFoundationsPopulationbasedSHM2021,gardnerFoundationsPopulationbasedSHM2021,tsialiamanisFoundationsPopulationbasedSHM2021}.
\subsection{Transfer Learning}
In PBSHM, one of the most widely-explored areas of transfer learning is known as domain adaptation in which feature data are mapped from a label-rich source domain to a label-scarce target domain, with the aim of reducing the distance between the source and target domain in a shared latent space such that label-information can be shared. Fink et al. discuss domain adaptation's place in fleet prognostics and health management \cite{finkPotentialChallengesFuture2020}. Gardner et al. utilised domain adaptation to transfer inferences across different structures, considering a population of laboratory-scale three-story buildings \cite{gardnerApplicationDomainAdaptation2020}. Zhang et al. used domain adaptation for fault diagnosis in the context of rotating machinery and between different machines \cite{zhangNewDeepLearning2017,liIntelligentCrossmachineFault2020}. Xu et al. showed that transfer learning can be used to diagnose story-wise damage conditions of buildings effected by earthquakes \cite{xuPhyMDANPhysicsinformedKnowledge2021}.
\subsection{Active Learning} 
Active learning (AL) presents another avenue to address the limitations imposed by data scarcity in SHM. Conventional supervised machine-learning methods are infeasible for many SHM applications because of the costs associated with descriptive labels. This has led to the development of unsupervised and semi-supervised machine-learning techniques. Active learning is a set of techniques that selectively queries labels for otherwise unlabelled data that are most informative given the current model; the model can then be updated using this informed subset of labelled data. Active learning can be applied offline to a large pool of collected data \cite{wangActiveLearningDensity2017}, or online, whereby the dataset is continuously updated as new observations are collected \cite{zhuActiveLearningData2007}. The online setting is particularly significant in SHM; generally data from a monitored structure will become available gradually throughout the life of the structure. In many cases, inspecting monitored systems can be extremely costly, so if a system can determine when only the most critical or informative observations need to be investigated, this could lead to significant reductions in maintenance costs \cite{bullProbabilisticFrameworkOnline2019}.

Active learning has seen growing interest, particularly within SHM; Bull et al. provide an online active learning framework for the classification problem and show the effects of the framework via a case study on acoustic emissions data \cite{bullProbabilisticFrameworkOnline2019}. Hughes et al. present a risk-based formulation of active learning in which queries are guided by the expected value of information \cite{hughesRiskbasedActiveLearning2022} and outline a method to minimise the effects of sampling bias in active learning \cite{hughesRobustRiskbasedActivelearning2022}.

In the literature, most research on active learning focusses on the classification setting. However, for many real-life applications, such as degradation \cite{shahrakiReviewDegradationModelling2017} and crack growth \cite{heParameterEstimationMultiaxial2023}, regression-based models are more suitable. The work done in the area of active learning for regression (ALR) is comparatively much less. One of the first statistical analyses of ALR was provided by \cite{cohnActiveLearningStatistical1996}, where a locally-weighted regression model for studying the dynamics of a robot arm is showcased. More recently, \cite{wuActiveLearningRegression2019} proposed a 'greedy sampling' method for active learning regression. In that paper a greedy sampling method proposed by \cite{yuPassiveSamplingRegression2010}, is expanded to be 'greedy' in the output space and is shown to be a robust and effective method for active learning. \cite{caiMaximizingExpectedModel2013} proposed an expected model change maximisation framework for regression, the learner chooses the unlabelled instance which causes the maximum change in the current model parameter. \cite{freundSelectiveSamplingUsing1997} showed that query-by-committee is not only applicable to binary labels but also to discrete labels. For more thorough surveys of active learning the reader is directed to \cite{kumarActiveLearningQuery2020a,fuSurveyInstanceSelection2013,aggarwalActiveLearningSurvey2014}.

This article aims to combine the effects of the population-based approach with the efficiencies of active learning to improve decision-making, reduce costs and more effectively allocate resources for SHM. This model will be applied to a population of machining tools.

\subsection*{Novel Contribution}
We propose a novel approach to risk-based active learning for regression and combine it with information sharing via a hierarchical Bayesian model, where decisions are made to minimise the risk. Therefore, our contributions offer three practical advantages;
\begin{enumerate}
  \item Extending the value of measured data
  \item Extending the value of labels
  \item Relating both to a population-based decision analysis, which considers the data across a group of similar machines.
\end{enumerate}

\subsection*{Paper Outline}
Section \ref{sec:2} is organised into three subsections, the first shows a general framework for information sharing via hierarchical modeling, the second describes decision theory and computation of expected utility and risk, and the third subsection describes risk-based active learning for decision analysis. Section \ref{sec:3} introduces a case study and applies the frameworks for hierarchical modelling and decision theory from Section \ref{sec:2}.

\section{Methodology}
\label{sec:2}

The first part of this section outlines a framework for sharing information within homogeneous populations of structures which forms the basis of the probabilistic regression model used in the current case study. The second part includes an introduction to decision theory which forms the basis of the active-learning procedure.

\subsection{Hierarchical Bayesian Modelling}
\label{sec:2.1}

Section \ref{sec:2.1} follows an explanation provided by the authors precious work \cite{bullHierarchicalBayesianModeling2023a}. Consider data recorded from a population of K engineering structures. The population can be denoted,

\begin{equation}\label{eq:data}
  \left\{\mathbf{x}_k, \mathbf{y}_k\right\}_{k=1}^K=\left\{\left\{x_{i k}, y_{i k}\right\}_{i=1}^{N_k}\right\}_{k=1}^K
\end{equation}

\noindent where $y_{k}$ is a target response vector for inputs $x_{k}$ and $\{x_{ik},y_{ik}\}$ are the $i^{\text{th}}$ pair of observations in group $k$. There are $N_{k}$ observations in each group and thus $\Sigma^{K}_{k=1}N_k$ observations in total. The aim is to learn a set of $K$ predictors related to a regression or classification task. This paper focuses on regression, where the tasks satisfy,

\begin{equation}\label{eq:regression}
  \left\{y_{i k}=f_k\left(x_{i k}\right)+\epsilon_{k}\right\}_{k=1}^K
\end{equation}

\noindent and the output $y_{ik}$ is determined by evaluating one of $K$ latent functions with seperate additive noise $\epsilon_{k}$. The mapping $f_k$ is assumed to be correlated between members in the population. The models should be improved by learning the parameters in a joint inference over the whole population. In machine learning this is referred to as multi-task learning (MTL); in statistics, such data are usually modelled with hierarchical models \cite{gelmanDataAnalysisUsing2006,kreftIntroducingMultilevelModeling1998}.

In practice, some members in a population may possess extensive historical data, while members that may have been more recently deployed will have very limited data for training. Learning separate independent models for each group might lead to unreliable results for data-poor members, while a single regression model of all the data (complete pooling) would result in poor generalisation. Hierarchical Bayesian models can learn separate models for each member, while encouraging the parameters of these models to be correlated \cite{murphyMachineLearningProbabilistic2012}. The established theory is summarised here.

Consider K linear regression models,

\begin{equation}\label{eq:klinearregressions}
  \bigl\{\mathbf{y}_{k} = \mathbf{\Phi}_k\mathbf{\alpha}_k + \epsilon_k\bigr\}_{k=1}^K
\end{equation}

\noindent where $\mathbf{\Phi}_k = [\mathbf{1},\mathbf{x}_k]$ is the $N_k \times 2$ design matrix; $\mathbf{\alpha}_k$ is the $2 \times 1$ vector of weights; and the noise vector is $N_k \times 1$ and normally distributed $ \epsilon_k\sim N(0,{\sigma^{2}}_{k}\mathbf{1})$. $\mathbf{1}$ is a vector of ones, $\mathbf{I}$ is the identity matrix, and $N(m,s)$ is the normal distribution with mean $m$ and (co)variance $s$. The likelihood of the target response vector is then 

\begin{gather}\label{eq:likelihood}
  \mathbf{y}_k \mid \mathbf{x}_k \sim \mathcal{N}(\mathbf{\Phi}_k\mathbf{\alpha}_k,{\sigma_k}^{2}\mathbf{I}) \\
   \therefore y_k \mid x_k \sim \mathcal{N}({\alpha_1}^{(k)}+{\alpha_2}^{(k)}x_{ik},{\sigma_k}^{2})
\end{gather}

\noindent following the Bayesian methodology, one can set a common hierarchy of prior distributions over the weights (slope and intercept) for each member in the population, typically normal distributions are used for the weights of each group and inverse-Gamma for the variance parameter. 

\begin{equation}\label{eq:alpha prior}
  \left\{\mathbf{\alpha}_K\right\}_{k=1}^K \stackrel{i.i.d.}{\sim} \mathcal{N} \left(\mathbf{\mu}_{\alpha},\operatorname{diag}\left\{\mathbf{\sigma}^2_\alpha\right\}\right)
\end{equation}

\begin{equation}\label{eq:mu prior}
  {\mathbf{\mu}}_{\alpha} \sim \mathcal{N} \left(\mathbf{m}_{\alpha},\operatorname{diag}\left\{\mathbf{s}_\alpha\right\}\right)
\end{equation}

\begin{equation}\label{eq:sigma prior}
  {\mathbf{\sigma}}_{\alpha} \stackrel{i.i.d.}{\sim} \mathcal{IG} \left(a,b\right)
\end{equation}

\begin{figure}[h!]
  \includegraphics[width=0.85\textwidth]{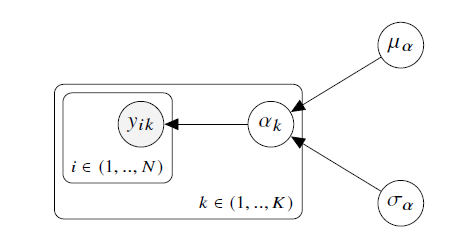}
  \caption{A graphical model representing the linear mixed model with partial pooling.}
  \label{fig:dgm}
\end{figure}

In words, Equation \ref{eq:alpha prior}  assumes that the weights  $\left\{\mathbf{\alpha}_K\right\}_{k=1}^K$ are normally distributed $N(\cdot)$ with mean $\mu_\alpha$ and covariance $\operatorname{diag}\left\{{\sigma^2}_{\alpha}\right\}$. Equation \ref{eq:mu prior} states that prior expectation of the weights $\alpha_k$ is normally distributed with mean $\mu_\alpha$ and covariance $diag\left\{\mathbf{s}_\alpha\right\}$. Equation \ref{eq:sigma prior}  states that the prior deviation of the slope and intercept is inverse-Gamma distributed with shape $a$ and scale $b$. A general representation of hierarchical regression can be seen in the direct graphical model in Figure \ref{fig:dgm}.

 %Selecting appropriate hyperparameters $\left\{\mathbf{m}_\alpha,\mathbf{s}_\alpha,a,b\right\}$, is critical to the performance of hierarchical models

The parent nodes $\left\{\mathbf{\mu}_\alpha,{\mathbf{\sigma}^2}_\alpha\right\}$ are inferred from the data so Equations \eqref{eq:alpha prior}--\eqref{eq:sigma prior} encode prior belief of the dependence between latent variables. If these parent nodes were a fixed value, rather than inferred, each model would be independent, preventing the flow of information between domains, this structure allows data-sparse domains to borrow statistical strength from those that are data rich \cite{murphyMachineLearningProbabilistic2012}. 

\cite{huangHierarchicalSparseBayesian2015,huangMultitaskSparseBayesian2019} present an early example of using Hierarchical Bayesian models to represent engineering structures for SHM. \cite{bullHierarchicalBayesianModeling2023a} used hierarchical models to improve the survival analysis of a truck fleet and power prediction in a wind farm. \cite{difrancescoDecisiontheoreticInspectionPlanning2021} used hierarchical models to account for incomplete and imperfect data in an inspection-planning setting.

\subsection{Decision Theory}\label{sec:2.2}

Hierarchical models allow one to quantify beliefs about the states of interest and do reasoning under uncertainty. In engineering these predictions are used to take actions in the real world. Decision theory gives a framework to use the uncertainty quantification provided by a Bayesian approach to make optimal choices in many engineering problem settings. 

The idea of a 'rational' decision-maker is defined as a decision-maker that acts to maximise the expected-utility of their actions; this is expressed via the von Neumann-Morgenstern theorem \cite{morgensternTheoryGamesEconomic1964}, which states that, for two decidable actions $a$ and $b$:

\begin{equation}\label{eq:rational}
  a \succeq b \iff EU(a) \geq EU(b)
\end{equation}

\noindent where $\succeq$ denotes a weak preference, indicating that a decision-maker favours $a$ at least as much $b$; and $EU(\cdot)$ denotes the expected utility associated with doing an action. In \cite{morgensternTheoryGamesEconomic1964}, the expected utility is derived. Consider a stochastic event $X$. Which has mutually-exclusive outcomes of which $x \in \mathcal{X}$ are conditionally dependent on a decision $D$ between actions $a$ and $b$. The expected utility of action $a$ is computed as follows,

\begin{equation}\label{eq:EU}
  EU(a) = \sum_{x \in \mathcal{X}} P(X = x \mid D = a) \cdot U(X = x,D = a)
\end{equation}

\noindent $P(X \mid D=a)$ is the probability of the outcome of $X$ given that action $a$ is executed. $U$ denotes a utility function that maps $U: \mathcal{X} \times \mathcal{A} \rightarrow \mathbb{R}$. Conveniently, if the utility of $D$ is independent of the variable X $U(X,D)$ can be expressed as the sum of two utility functions, $U(X)$ and $U(D)$ That separately describe the utilities associated with outcomes and actions. Equation \ref{eq:EU} can then be written as,

\begin{equation}\label{eq:UDUa}
  EU(a) = \biggl[\sum_{x \in \mathcal{X}} P(X = x \mid D = a) \cdot U(X = x)\biggr] +U(D=a)
\end{equation}

For a single decision $D$ over a finite set of actions $\mathcal{A}$, an optimal action $a^*$ can be defined such that the maximum expected utility (MEU) is achieved, where,
 
\begin{equation}\label{eq:MEU}
  \mathrm{MEU}(D) = \max_{a \in \mathcal{A}}EU(a)
\end{equation}

\noindent and,

\begin{equation}\label{eq:a*}
  a^* = \underset{a \in \mathcal{A}}{\arg\max}EU(a)
\end{equation}

From Equations \ref{eq:UDUa} and \ref{eq:MEU}, one can see an equivalence between expected utility and risk, which is defined as the product of a probability and a cost. One limitation of utility/risk-based decision theory is that it can be difficult to obtain the probability distributions in Equations \ref{eq:EU} and \ref{eq:UDUa}. However, via the Bayesian framework outlined in Section \ref{sec:2.1} one can acquire posterior distributions, based on one's beliefs about an action, that can be used as the probabilities required for Equations \ref{eq:MEU} and \ref{eq:UDUa}. Additionally, the costs or utilities required for these equations can be elicited from asset-owners, allowing the expected utility of an action to be estimated.

\subsection{Active learning}\label{sec:2.3}
 
To implement a Bayesian hierarchical model, such as the one proposed in Section \ref{sec:2.1}, labelled data is required. As discussed in Section \ref{sec:1}, for many engineering applications, and particularly for SHM, collecting labels for data is very expensive. It requires sending a domain expert to inspect the physical asset and often requires the operation of the asset to be temporarily halted. These costs motivate reducing the number of inspections and only inspecting measurements that would most improve the predictive model. Active learning is a set of tools that do this. Additionally, an offline active learner would not be suitable for SHM. SHM systems are required to analyse data as it arrives throughout the life of the monitored system and decisions about interventions can be required immediately upon arrival of this data. One of the major challenges of an \emph{online} active learner is that once the decision is made \emph{not} to query a label, access is lost to this information. It is not possible to retrieve the label once the opportunity has passed.    

Consider data, $X = {\left\{\mathbf{x}_i\right\}^N}_{i=1}$, which have hidden labels, $Y = {\left\{\mathbf{y}_i\right\}^N}_{i=1}$, which can be acquired by paying the costs associated with inspection. The process of choosing and labelling these data points is referred to as \emph{querying}. An active learner aims to learn a mapping of the observations, $X$, to the labels, $Y$, while keeping queries to a minimum. A general heuristic for active learning is presented by \cite{bullProbabilisticActiveLearning2019a} and adapted to regression below,

\begin{figure}[h] 
  \centering
  \includegraphics[width=0.85\textwidth]{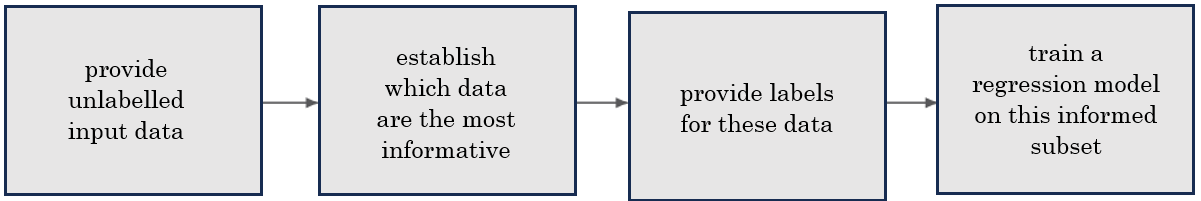} 
  \caption{An active learning heuristic adapted from \cite{bullProbabilisticActiveLearning2019a}}
  \label{fig:AL_hueristic}
\end{figure}

The measure that an active learner uses to decide which unlabelled data to query is important. Information theoretic approaches use information measures such as entropy and uncertainty to guide querying, these types of approaches are common. \cite{hughesRiskbasedActiveLearning2022} suggest a risk-based active learner for SHM. This approach uses the expected utility of an action, or the 'Value of Information', to guide querying. While risk-based active learning has been explored in the classification setting \cite{hughesRobustRiskbasedActivelearning2022}, the current paper proposes an approach to risk-based active learning for the regression problem in which queries are guided according to the expected utility.

\section{Case Study - A Population of Machining Tools} \label{sec:3}

In this section, the framework outlined in Section \ref{sec:2} will be applied to a case study. A population of machining tools will be modelled by a hierarchical Bayesian model and a risk-based decision process will be used to actively inspect the tools. The decision-theoretic approach to inspection planning will be compared to a periodic inspection plan.

\subsection{The dataset}\label{sec:3.1}

\begin{figure}[h!] 
  \centering{\includegraphics[scale=0.45]{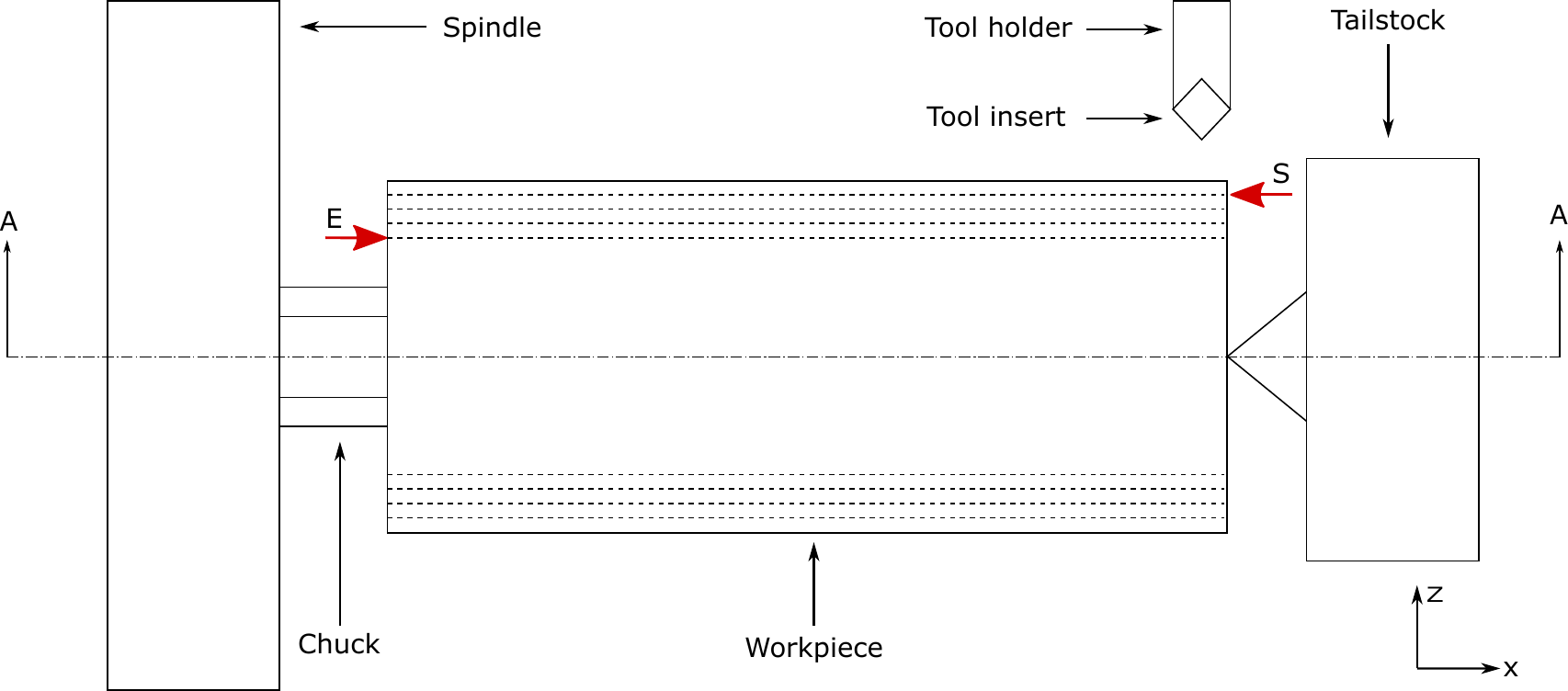}}
    \caption{Schematic showing the experimental set-up used for data acquisition  \cite{wickramarachchiAutomatedTestingAdvanced2019a}
  }
\label{fig:experiment}
\end{figure}

A dataset described in the authors previous work \cite{wickramarachchiAutomatedTestingAdvanced2019a} measures deterioration over the life of machining tools during a turning process.  The experimental set-up is shown in Figure \ref{fig:experiment}. The workpiece is rotated around the dashed A-line and the tool makes four passes along the workpiece. Each pass starts at point S and ends at point E. After four passes, the tool is inspected, and measurements are taken of the workpiece and tool. This process continued until tool failure. The investigation was repeated for seven nominally identical tools, which form a population.

\begin{figure}[!h] 
  \centering{\includegraphics[scale=0.8]{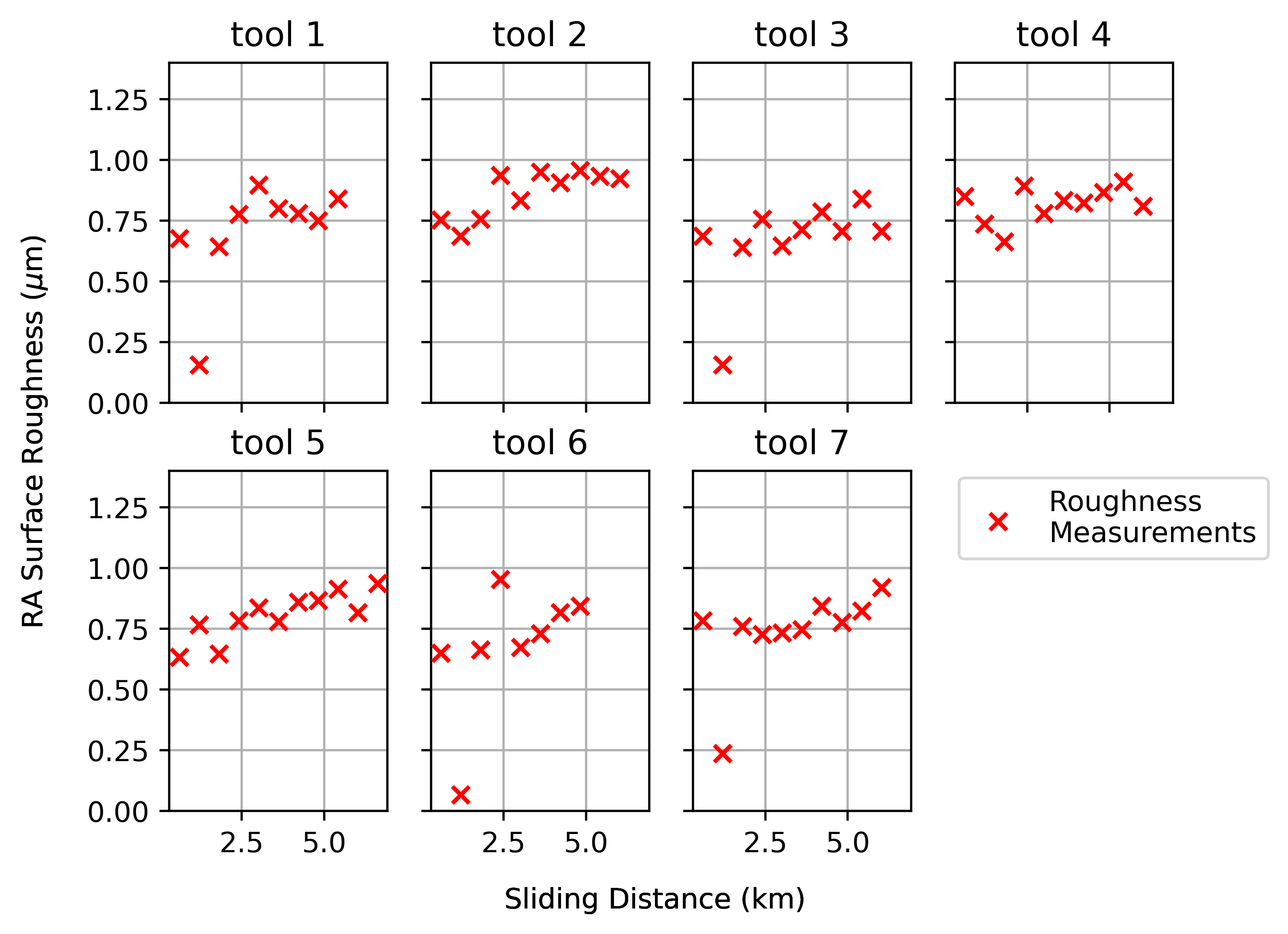}}
    \caption{Experimental surface roughness measurements}
\label{fig:data}%
\end{figure}

The deterioration is measured indirectly from the roughness of the workpiece. As the machining tool deteriorates, the surface quality of the workpiece will deteriorate and the roughness will increase. The measurements from this experiment can be seen in Figure \ref{fig:data}. It can be seen that, in general, surface roughness increases with the distance a given tool has cut along the workpiece, this distance is termed \emph{sliding distance}. Several tools show a sharp decrease in surface roughness at the second measurement point --- this reading is believed to occur due to an initial sharpening of the tool early in the cutting process.

% when introducing dataset, talk about shallow gradient, noisy measurements, linear regression best

Because of the nature of the experiment, the measurements of surface roughness are very noisy; which can lead to robustness issues when modelling. Combined with the high noise, the shallow gradient of the deterioration makes it difficult to learn the parameters of the regression. This motivates the use of a Bayesian hierarchical model adapted from Section \ref{sec:2.1}.

\subsection{The hierarchical model}\label{sec:3.2}

\begin{figure}[h!]
 \includegraphics[width=0.85\textwidth]{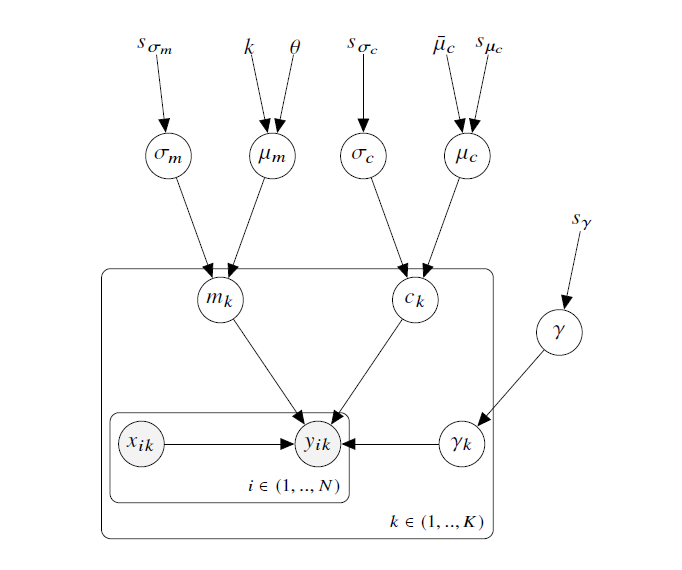} 
  \caption{A graphical model representing the linear mixed model with partial pooling}
  \label{fig:hierarchical_dgm}
\end{figure}

Because of the natural degradation of tools during the machining process, and its effect on surface finish, tools must be replaced regularly. While each tool may be produced to the same specification and made from the same materials, there will be variations in the manufacturing process that manifest as variations between the physical properties of the tools and differences in behaviour between the tools; this can be an issue for standard modelling techniques. However, this variation lends itself well to a hierarchical Bayesian model because these types of model account for variations within a population while taking advantage of the statistical similarities between them. An additional benefit of hierarchical models is their suitability to the online setting and sparse datasets. This is particularly useful for tool condition monitoring where researchers may need to make predictions as soon as the machining process has begun and with only a few data points to learn a model. Additionally, the usual benefits of Bayesian modelling apply (uncertainty quantification, prior information etc.).
The hierarchical framework set out in Section \ref{sec:2.1} is adapted below to the case study. Here, there is a population of $K=7$ similar tools, with a target response vector $y_{k}$, the roughness measurements for each tool. The input vectors $x_{k}$ are the sliding distance measurements for each tool (how far the tool has cut across the work piece). $\{x_{ik},y_{ik}\}$ are the $i^{\text{th}}$ pair of observations in tool $k$. There are $N_{k}$ observations in each member and thus $\Sigma^{K}_{k=1}N_k$ observations in total. The aim is to learn a set of $K$ predictors related to the regression task. The type of hierarchical model used for this analysis is a linear mixed model \cite{kreftIntroducingMultilevelModeling1998}, so for each member in the population, a gradient $m_k$ and intercept $c_k$ are learnt. A graphical representation of this model can be seen in Figure \ref{fig:hierarchical_dgm}.

The likelihood of this model is

\begin{equation} \label{eq:likelihood}
  \left\{{y}_{ik}\right\}_{k=1}^K {\sim} \mathrm{Cauchy}\left(y_{mean},{\gamma}_{k}\right)
\end{equation}

\noindent where the location parameter is the equation of a straight line,

\begin{equation}\label{eq:straight_line}
  y_{mean} = {m}_{k}\cdot{x}_{ik}+{c}_{k}
\end{equation}

\noindent The Cauchy distribution was chosen as the likelihood because of the noisy nature of the data. Cauchy distributions are particularly suited to these types of measurements because of the larger probability density at the extremes, compared to normal distributions another more typical choice. This makes the model less susceptible to outliers and extreme values. Following the Bayesian methodology, one can set prior distributions over the slope for the groups, which encode our prior knowledge of the parameter values.

\begin{equation} \label{eq:m_k}
  \left\{{m}_k\right\}_{k=1}^K {\sim} \mathrm{Normal}\left({\mu}_m,{\sigma}_{m}\right)
\end{equation}

\begin{equation} \label{eq:mu_m}
  {\mu}_m \sim \mathrm{Gamma}\left({k,\theta}\right)
\end{equation}

\begin{equation} \label{eq:sig_m}
  {\sigma}_m \sim \mathrm{HalfCauchy}\left(s_{\sigma_{m}}\right)
\end{equation}

\noindent Where the slopes are $\mathrm{normally}$ distibuted, with mean ${\mu}_m$ and standard deviation $\sigma_{m}$. Equation \ref{eq:mu_m} shows the prior expectation of the slopes is $\mathrm{Gamma}$ distributed with shape $k=1$ and scale $\theta=1$. The Gamma distribution was a natural choice because we want to encode that the gradients are always positive over the life of a tool. Equation \ref{eq:sig_m} shows that the prior deviation of the slope is $\mathrm{HalfCauchy}$ distributed with location parameter equal to zero and scale parameter $s_{\sigma_{m}}=25$. As recommended by \cite{gelmanPriorDistributionsVariance2006a}, the variance priors for this hierarchical model are set to be \emph{weakly informative}. The priors for the intercepts are

\begin{equation} \label{c_k}
  \left\{{c}_k\right\}_{k=1}^K {\sim} \mathrm{Normal}\left({\mu}_c,{\sigma}_{c}\right)
\end{equation}
\begin{equation} \label{eq:mu_c}
  {\mu}_c \sim \mathrm{Normal}\left({\bar{\mu}}_{c},s_{\mu_{c}}\right)
\end{equation}
\begin{equation} \label{eq:sig_c}
  {\sigma}_c \sim \mathrm{HalfCauchy}\left(s_{\sigma_{c}}\right)
\end{equation}
where the intercepts are $\mathrm{normally}$ distributed, with mean ${\mu}_c$ and standard deviation $\sigma_{c}$. Equation \ref{eq:mu_c} shows the prior expectation of the intercepts is also $\mathrm{normally}$ distributed with mean $\bar{\mu_{c}}=0$ and standard deviation $s_{\mu_{c}}=1$. Equation \ref{eq:sig_c} shows that the prior deviation of the intercept is $\mathrm{HalfCauchy}$ distributed with location parameter equal to zero and scale parameter $s_{\sigma_{c}}=25$.

The systematic application of graph-theoretic algorithms has led to a number of probabilistic programming languages. Here, models are implemented in \texttt{NumPyro} \cite{phanComposableEffectsFlexible2019c}. The parameters are inferred using MCMC, via the no U-turn implementation of Hamiltonian Monte Carlo \cite{hoffmanNoUTurnSamplerAdaptively2014}. Throughout, the burn-in period is 1000 iterations and 2000 iterations are used for inference.

To motivate a population-based approach, the hierarchical model will be compared to a model with complete-pooling and an independent model with no pooling at all. In the complete-pooling approach, a single regression for all the data is learnt. For the independent model, a regression is learnt for each member in the population but it is assumed there is no correlation between  them. Hierarchical modelling is somewhere in between, where a separate regression can be learnt for each member, while encouraging the parameters of these models to be correlated \cite{murphyMachineLearningProbabilistic2012}.

To simulate tool replacement, some measurements from Figure \ref{fig:data} will be hidden from the models, this emulates new tools with scarce data.

\begin{figure}
  \centering
  \includegraphics[width=0.85\textwidth]{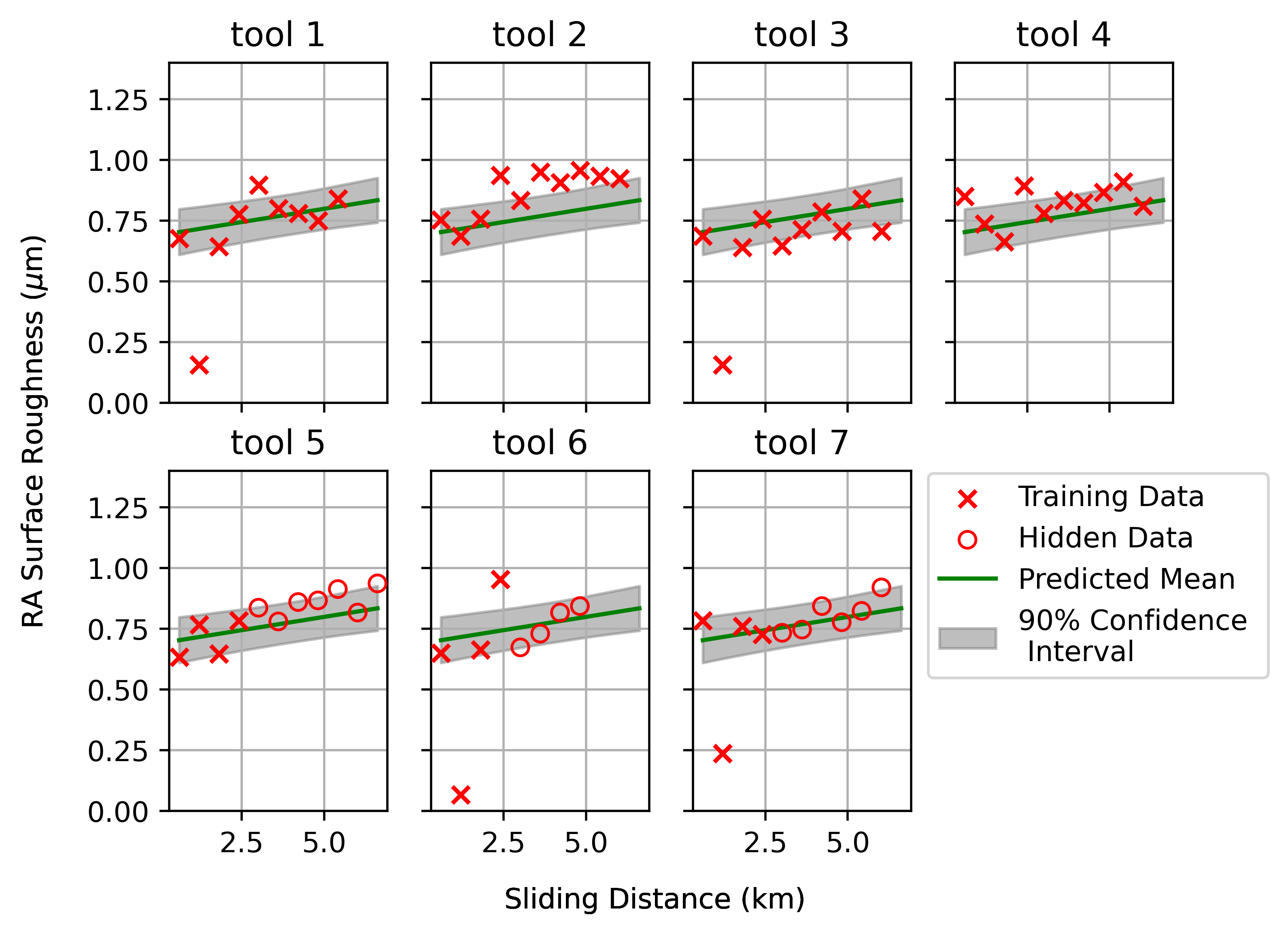}
  \caption{Predictions using a complete-pooling method}
  \label{fig:complete_pooling}
\end{figure}

Figure \ref{fig:complete_pooling} shows that with complete-pooling, the model struggles with poor generalisation and the physical variations between tools means that a single regression performs poorly on many tools. Even with access to a tools full history, such as Tool 2 and Tool 3, a complete-pooling method may not perform well.

\begin{figure}
  \centering
  \includegraphics[width=0.85\textwidth]{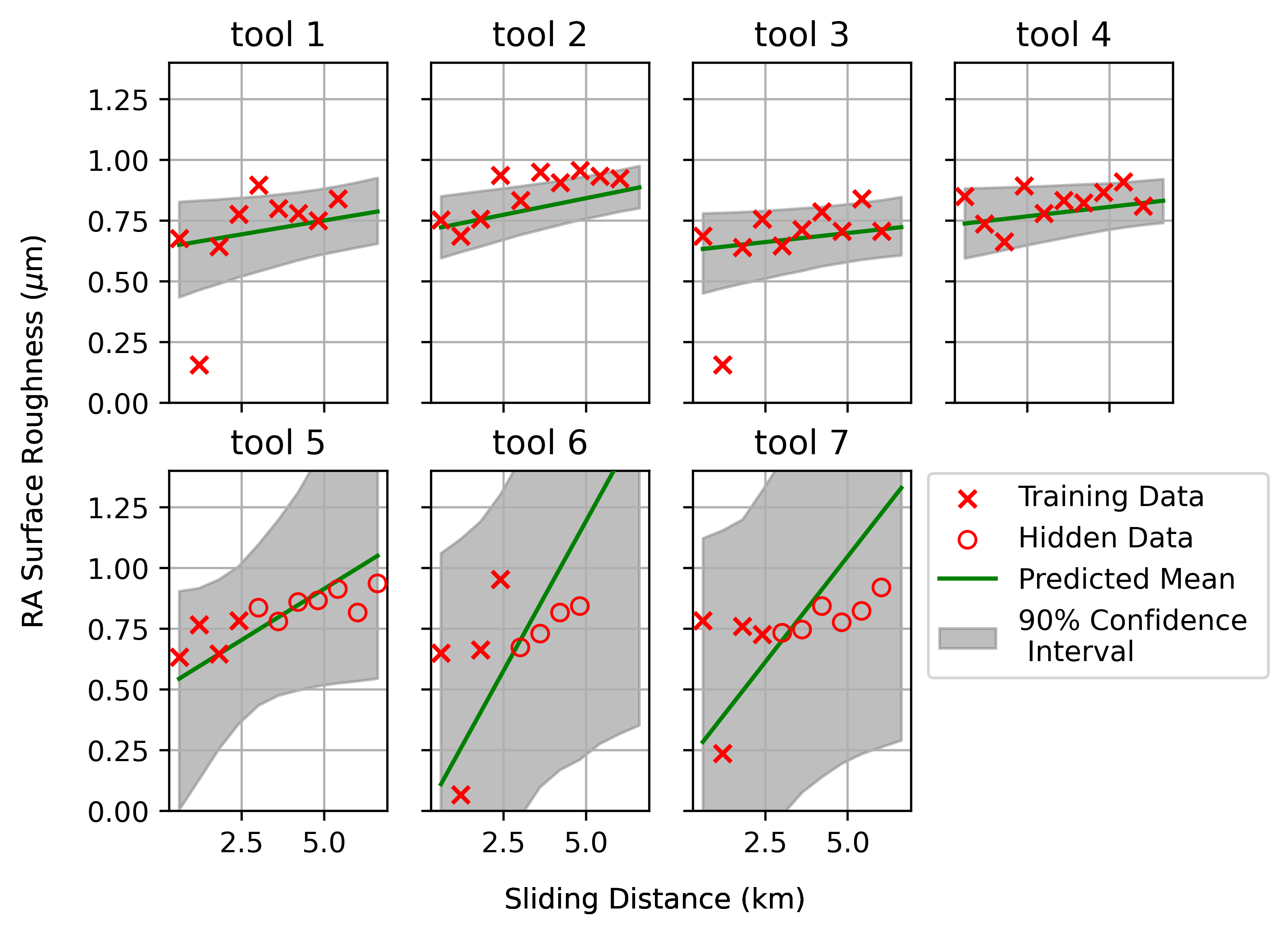}
  \caption{Predictions without any pooling method}
  \label{fig:no_pooling}
\end{figure}

The model with no pooling, can be seen in Figure \ref{fig:no_pooling}. For tools that have enough historic data, the mean and variance predictions are reasonable. However, this model performs poorly with scarce data. For these tools, the model struggles to learn the parameters of the regression. The variance of these tools are over-estimated and makes poor predictions about the hidden data, when there is not enough data the model relies on vague priors. Over predicting the variance of the surface roughness could be problematic for asset owners if they use this model to inform decisions. For example, unnecessary inspections may be triggered or tools may be replaced prematurely, increasing the costs of production. 

\begin{figure}
  \centering
  \includegraphics[width=0.85\textwidth]{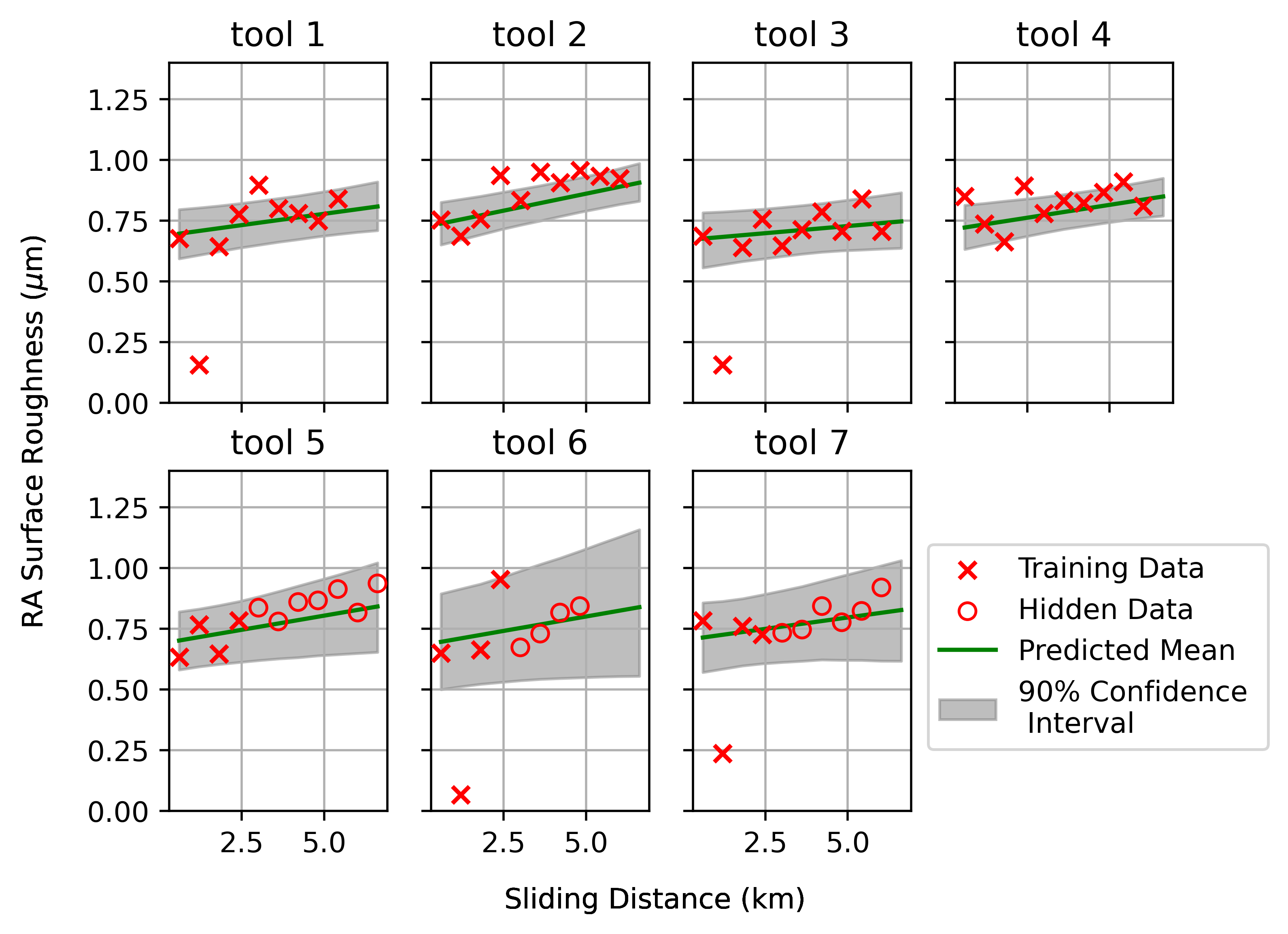}
  \caption{Predictions using a partial pooling method}
  \label{fig:partial_pooling}
\end{figure}

Finally, the hierarchical model can be seen in Figure \ref{fig:partial_pooling}. For tools with plentiful data, this model behaves comparably to the no-pooling model. However, for data-scarce tools, there is a large reduction in the estimated variation and improvements in the predicted mean. This model is able to draw on the statistical strength of other tools to help predictions. The model \textit{remembers} the data from other tools, captured by the prior shared between models of machining operations, and has learnt how similar tools behave.

It can be seen in Table 1 that the information sharing provided by a partial-pooling model can improve regression accuracy, shown through a reduction in the total mean squared error as compared to the complete-pooling and no-pooling models.

\begin{table}[h!]
    \includegraphics[width=1\textwidth]{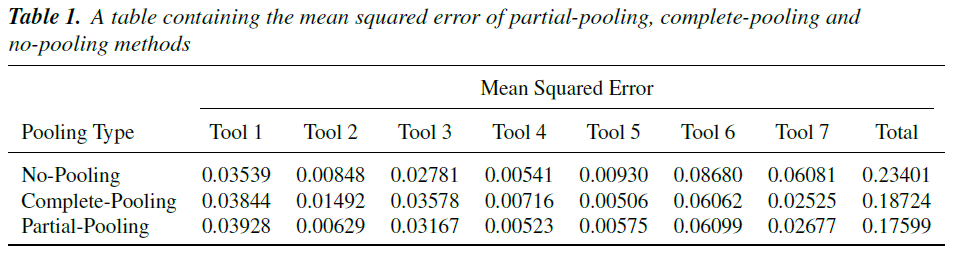} 
\end{table}

These results are in accordance with the results of the authors previous work \cite{bullHierarchicalBayesianModeling2023a,dardenoHierarchicalBayesianModelling2024} and motivates sharing information across a population to improve predictions.

\subsection{Decision Process and Active Learning}\label{sec:3.3}

\noindent Tool health deterioration can lead to tool failure. When monitoring tools, such as the ones that produced the data in this case study, a manufacturer may have an interest in avoiding tool failure because of the safety implications and costs associated with damage to the workpiece. Typically, inspections are used to observe the health state of the tool. However, performing an inspection has its own costs. Tool inspections halt production and labour and expertise are required to conduct an inspection. Optimising this decision process, whether or not to inspect an asset, can improve economic efficiency. Often there is a limited budget for inspections and ideally inspections should only be conducted when necessary. Employing decision-theoretic approach as shown in Section \ref{sec:2.2} to guide decision-making for inspections provides a risk-based active learning approach to more efficiently allocate inspection budgets.

A common quality-control criterion for the machining of engineering components is maintaining a surface roughness below some threshold level $S_{\text{crit}}$, beyond which the component is no longer fit for purpose. A high-quality surface finish can significantly improve the fatigue strength, corrosion resistance and creep life of machined parts \cite{sharkawySurfaceRoughnessPrediction2014a}. If the surface finish is damaged via inadequate modelling or control of the machining process, the part may need to be discarded or re-machined; this has an associated utility, $C_{\text{workpiece}}$. Additionally, during the machining process an inspection can be conducted to gain access to a noisy observation of the surface roughness, with a utility $C_{\text{inspection}}$. Throughout the machining process the tool can be replaced, with a cost $C_{\text{tool}}$. Here, there is a decision, inaction, to allow the tool to continue machining with a risk to damage the workpiece, and action, to inspect/replace the tool.

The EU, as seen in Equation \ref{eq:EU}, needs to be determined for each of the possible actions, and the action with maximum expected utility should be chosen in accordance with Equation \ref{eq:a*}. One limitation of the dataset used in this case study is that inspections and tool replacements can only be triggered at specific discrete time steps because the data was collected at periodic intervals, a dataset with less restrictive inspections points is under production and will be part of future work.

The expected utility of three actions need to be considered to evaluate this decision process. Inaction, to do nothing and allow the tool to continue machining, to inspect the tool and/or to replace the tool. If an inspection occurs, there is another opportunity to decide to replace the tool based on the new information acquired from the inspection.

% In the second layer of the decision tree another decision needs to be made. For this case study, only the first decision in Figure \ref{fig:decision_tree} will be analysed. After an inspection is triggered a second decision is required to decide between replacing the tool or continuing machining, as the dataset has already been collected and tool replacements can't be mandated retroactively, this second decision is not included in the current work. 

Again, Equation \ref*{eq:EU} can be used to calculate the expected utility of a decision. To compute the utility associated with inaction, \emph{not} inspecting the tool at time step $t$, one can use Equation \ref{eq:do_nothing}. The two outcomes of inaction are the surface roughness reaching or exceeding $S_{\text{crit}}$ before the next opportunity to intervene, time step $t+1$, or the surface roughness \emph{not} reaching or exceeding $S_{\text{crit}}$. The utility of not exceeding $S_{\text{crit}}$ is assigned a value of zero, so the second term in Equation \ref{eq:do_nothing} equals zero. The probability $P_{t+1}(S>S_{\text{crit}})$ can be estimated from the model; here it is the proportion of samples from the Hamiltonian Monte Carlo simulation that exceed $S_{\text{crit}}$ before the next opportunity to inspect. 

\begin{equation}\label{eq:do_nothing}
  EU(D = \text{do nothing}) = P_{t+1}(S > S_{crit}) \times U(S > S_{crit}) + P(S < S_{crit}) \times U(S < S_{crit})
  % + P(S>S_{crit}) \times P(t \leq t_f \geq t+1) \times U(S>S_{crit})
\end{equation}

When computing the utility of inspection at a given time step, again there are two outcomes based on whether the surface roughness will exceed $S_{crit}$. Again, probabilities of the outcomes can be estimated from the HMC samples. The equation to compute the EU of inspection can be seen in Equation \ref{eq:EU_inspection}. It includes the probability that the tool needs to be replaced, which is the probability that $S$ is already greater than $S_{\text{crit}}$ at that time step, as well as the probability that $S$ is \emph{not} greater than $S_{\text{crit}}$ but \emph{will be} by the next time step. For the current linear model, this is equivalent to $P_{t+1}(
S>S_{\text{crit}})$ and could be evaluated a such from the HMC samples. Again, these probabilities are estimated from HMC samples.    

\begin{equation}\label{eq:EU_inspection}
  EU(D = \text{inspection}) = C_{tool} \times (P_{t+1}(S>S_\text{crit})) + C_{inspection}
\end{equation}

The criteria required to trigger the replacement of a tool can be seen in Equation \ref{eq:tool_replacement_alpha}. It is the probability at which the risk associated with damaging the work piece (the probability of damage occurring multiplied by the utility associated with it) becomes greater than the cost of replacing the tool. 

\begin{equation}\label{eq:tool_replacement_alpha}
  \begin{aligned}
    P(T_{f}<T) \times C_{\text{workpiece}} &\geq C_{\text{tool}} \\
      P(T_{f}<T) &\geq \frac{C_{\text{tool}}}{C_{\text{workpiece}}} \\
    P(T_{f}<T) &\geq \alpha \\
 \end{aligned}
\end{equation}

Equation \ref{eq:tool_replacement_alpha} shows that tools should be replaced at the earliest time, T, that $P(T_{f}<T)$ is equal to or greater than the ratio, $\alpha$. $P(T_{f}<T)$ is the assessment of time-to-failure after one has updated beliefs with the results of any inspections and where failure is $S$ exceeding $S_{\text{crit}}$. Intuitively, as $C_{tool}$ increases, the probability requirements are increased and so replacements are more difficult to trigger, the system prioritises extending tool life. As $U(S>S_{crit})$ increases, the probability requirements are reduced and so replacements are more easily triggered and the system prioritises the surface quality of the workpiece.

A complete decision-theoretic approach would consider the expected value of information associated with improving the model based on the new data from an inspection. For the current work, it is assumed that inspecting the tool would provide zero improvement to the predictions of the model as the full value of information calculation is very computationally expensive.

\begin{figure*}[pt]
    \includegraphics[width=1\textwidth]{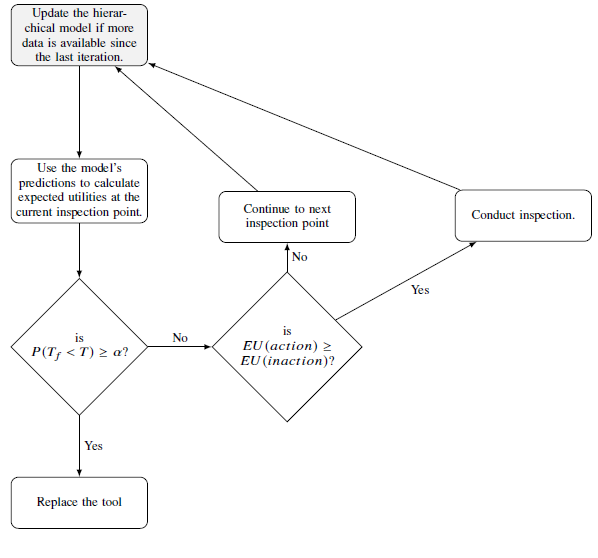} 
    \caption{The decision theoretic active-learning heuristic}
    \label{fig:heuristic}
\end{figure*}

At every potential inspection point, the hierarchical model makes predictions about the surface roughness. These predictions are based on the currently available data and are used to inform the decision analysis detailed above. A diagram showing this process is shown in Figure \ref{fig:heuristic}.

The online decision-theoretic active learning approach for inspection planning presented in this work is compared to a conventional inspection plan featuring periodic inspections. With both approaches, the model has access to the full measurement history of four of the tools and very limited data for the rest of the tools (the authors found that, in this case,  without access to at least four full tool datasets, the hierarchical model does not have enough data to draw on for accurate predictions).

In general, the parameters of the decision process, $S_{crit},C_{inspection},C_{tool},U(S>S_{crit})$ can be elicited from expert familiar with the inspection process. Here, the values are set by hand to reflect a situation common in industry, where the cost of damaging the workpiece is much greater than the cost of replacing the tool\footnote{It is worth noting that optimal actions are invariant under affine transformations of the utility function.}:

\begin{equation}\label{eq:decision parmeters}
  \begin{aligned}
    U(S>S_{crit}) &= 1 \\
    C_{tool} &= 0.25\\
    C_{inspection} &= 0.05 \\
    S_{crit} &= 0.9 \mu m
  \end{aligned}
\end{equation}

To assess the performance of these monitoring systems, each will be compared to a monitoring system with access to the full measurement history of every tool. This monitoring system will be seen as the gold standard as it represents the limiting case of using all possible information within the dataset and thus the point at which it decides to replace the tool can be considered to be the optimal point of replacement. If the other monitoring systems choose to replace the tool later than the optimal point, it is assumed the roughness has exceeded $S_{crit}$ (despite what the noisy measurements might suggest) and the workpiece will be considered damaged with a cost $C_{workpiece}$. If the monitoring systems choose to replace the tool earlier than the optimal replacement time, then some portion of the tool life is wasted, the cost of which can be calculated using Equation \ref{eq:early_inspection}. The optimal monitoring system can choose to replace the tool at any point throughout the life of the tools i.e. it can observe several measurements that are greater than $S_{\text{crit}}$ and then realise it should have replaced the tool before these measurements. The other monitoring systems are performing in an online manner, so do not have this luxury. The combined costs of both inspections and sub-optimal replacements, will be compared. 

\begin{equation}\label{eq:early_inspection}
  C_{\text{early replacement}} = \frac{t_{\text{replacement}}}{t_{\text{optimal replacement}}} \times C_{tool}
\end{equation}

\begin{figure}[h!] 
  \centering
  \includegraphics[width=0.85\textwidth]{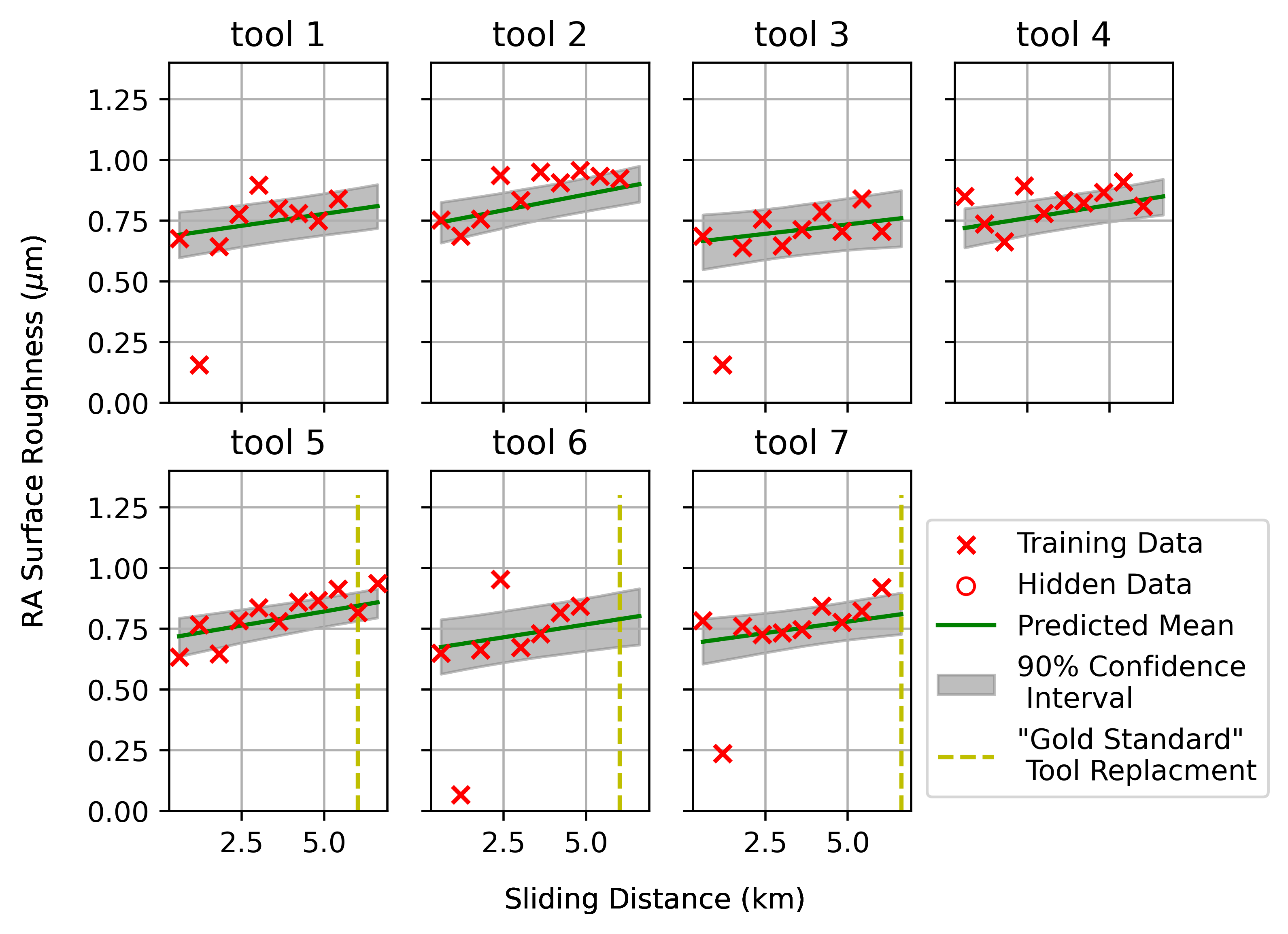}
  \caption{Benchmark replacements determined using all available information}
  \label{fig:replacement_all_data}
\end{figure}

As mentioned previously, the first four tools are to be seen as historic data that the manufacturer has collected from previous tools. The hierarchical model can leverage this data to inform the population-level distributions. The active-learning procedure tool replacements are not calculated for these tools, only Tools 5-7. During the original experiment the measurements were taken at a time step that relates to 6.02 km sliding distance \cite{wickramarachchiAutomatedTestingAdvanced2019a}. Figure \ref{fig:replacement_all_data} shows the optimal replacement for Tools 5-7 based on the "gold standard" model.

\section{Results}\label{sec:4}

In this section the decision theoretic active learner described in Section \ref{sec:3} and a periodic approach to inspection planning will be compared. The point at which these approaches suggest to replace Tools 5-7, based on the criteria showcased in Equation \ref{eq:tool_replacement_alpha}, will be compared to the "gold standard" model, for which, the suggested tool replacements are shown in Figure \ref{fig:replacement_all_data}.

\begin{figure}[h!] 
  \centering
  
  \includegraphics[width=0.85\textwidth]{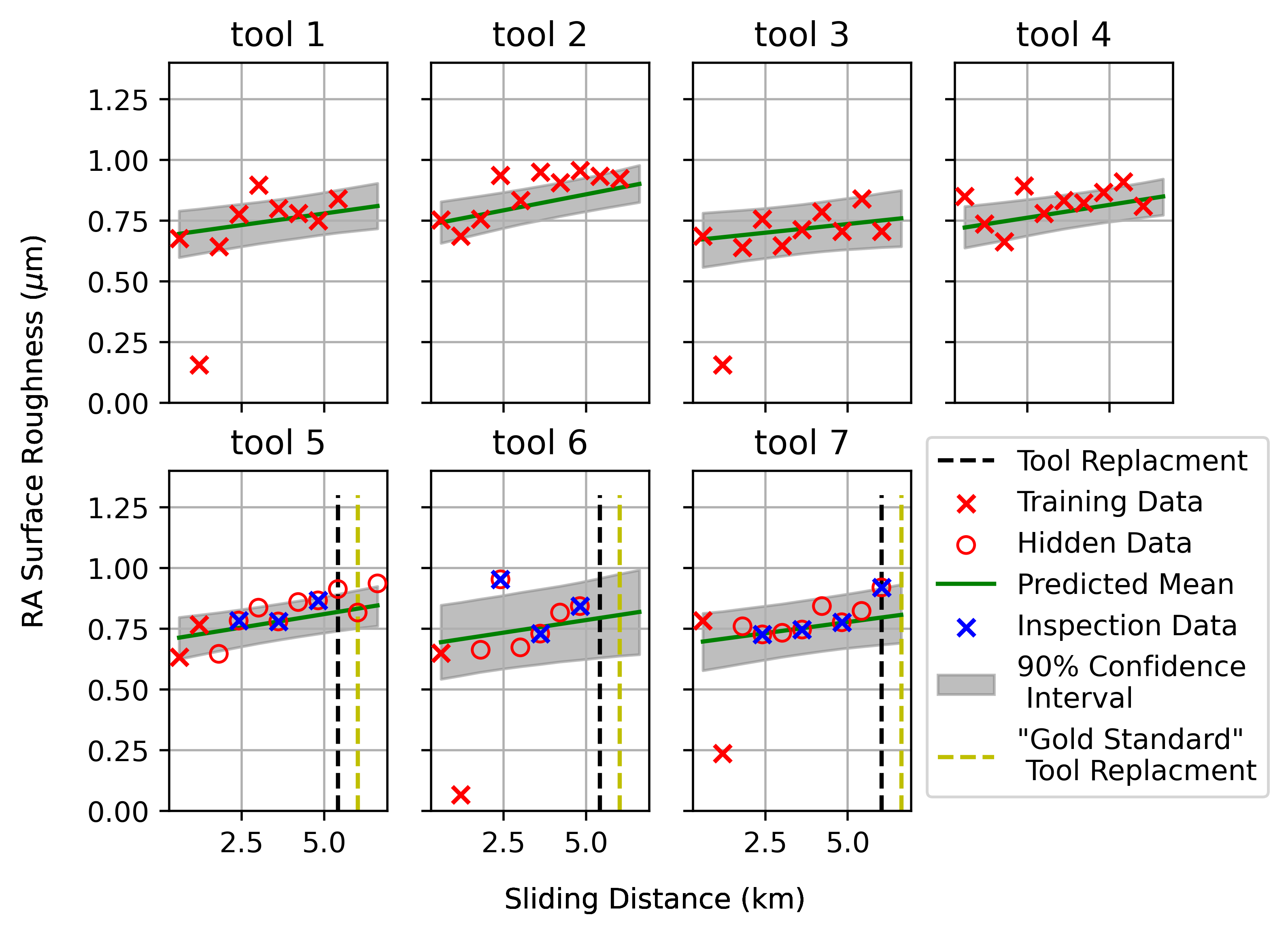}  
  \caption{Tool replacement with periodic inspections}
  \label{fig:replacement_periodic}
\end{figure}

Figure \ref{fig:replacement_periodic} shows the suggested tool replacements with periodic inspections. It can be seen that, compared to the optimal replacements in Figure \ref{fig:replacement_all_data}, every tool is replaced one time step too early, with a total of ten inspections. This result is to be expected because in forecasting the surface roughness predictions with reduced data (as compared to the "gold standard" model which has access to all the information), the uncertainty is inflated thus increasing the estimated risk associated with damage to the workpiece.  

\begin{figure}[h!] 
  \centering
  
  \includegraphics[width=0.85\textwidth]{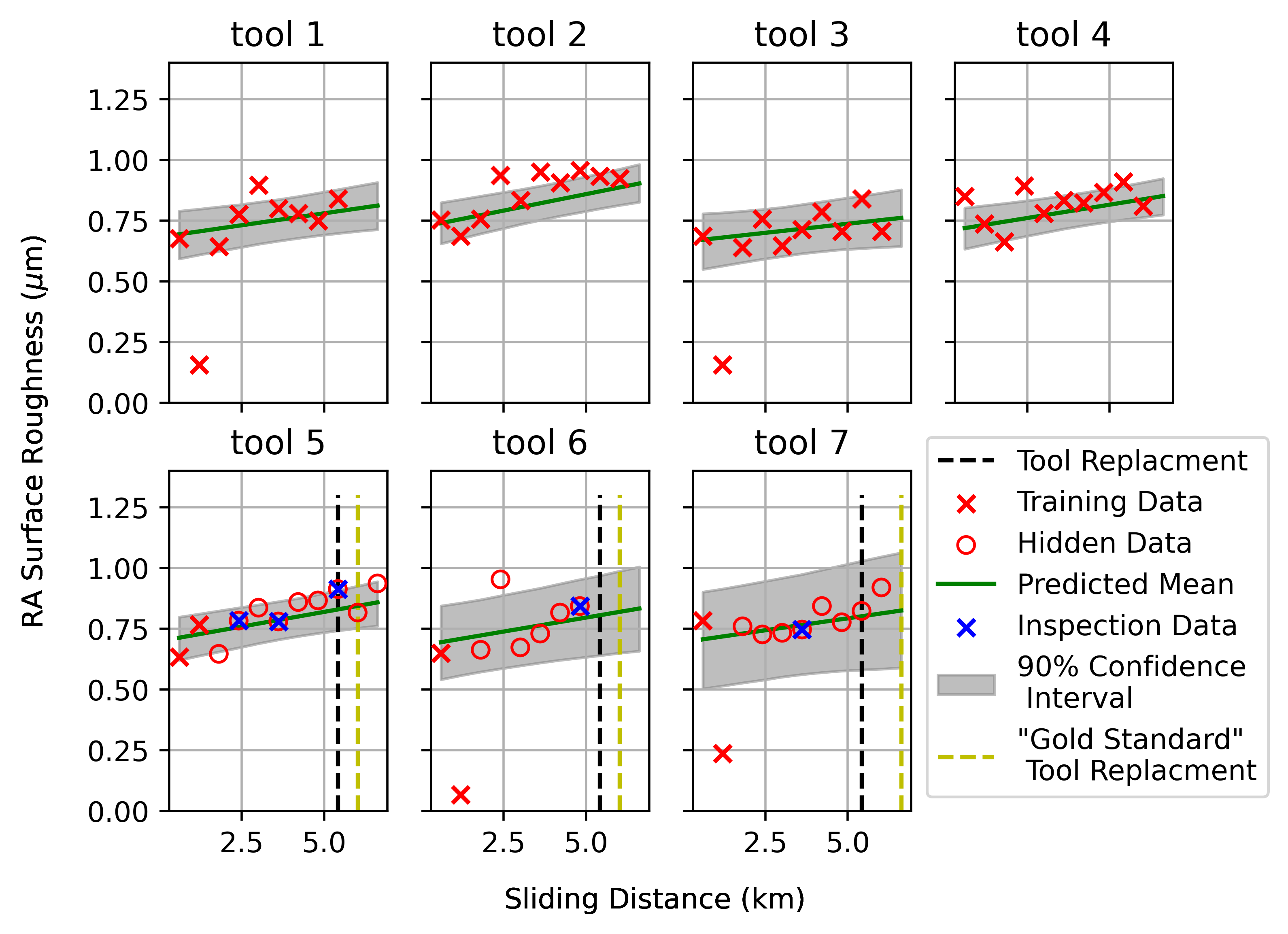}  
  \caption{Tool replacement with risk-based inspections}
  \label{fig:replacement_risk}
\end{figure}

Figure \ref{fig:replacement_risk} shows the replacements with risk-based inspection planning.  Tools 5 and 6 are again inspected one time step before the fully observed case. Additionally, Tool 7 is inspected two time steps before the gold standard. The risk-based monitoring system used a total of five inspections. The reduced number of inspections is because the decision-theoretic approach will, in general not suggest inspections early in the life of a tool when the risk of $S$ exceeding $S_{\text{crit}}$ is small. Most inspections will be triggered near the end of tool life when the risk associated with damaging the work piece is greatest. Again, because of the reduced access to information (compared to both the "gold standard" and the periodic inspections), inflated risk increases the likelihood of pre-optimal tool replacements.

Table 2 collates the number of inspections and tool replacements. It should be noted that the unit '0.602kms' refers to the sliding distance (how far the tool has cut) between each surface roughness measurement. 

\begin{table}[h!]
  \includegraphics[width=1\textwidth]{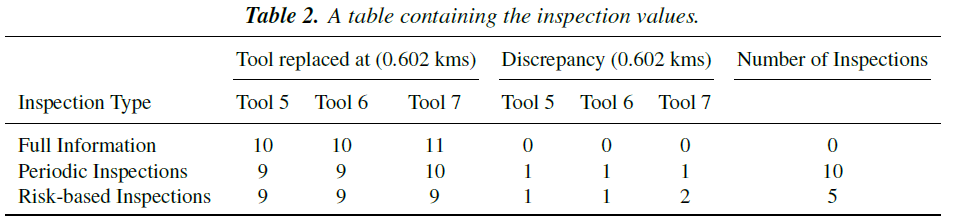} 
\end{table}

Table 3 compares the performance of the inspection approaches. The 'Cost of inspections' column is the number of inspections multiplied by the cost of an inspection. The 'Cost of Wasted Tool Life' for each inspection approach can be seen in the first column of Figure \ref{tab:costs} and can be calculated using Equation \ref{eq:tool_cost}.

\begin{equation}\label{eq:tool_cost}
  C_{wasted\ tool\ life} = C_{tool} \times \frac{\text{Discrepancy}}{\text{Optimal Replacement}}
\end{equation}

\begin{table}[h!]
\includegraphics[width=1\textwidth]{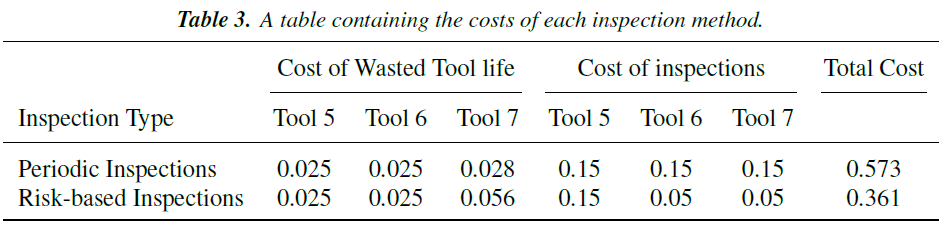} 
\end{table}

As can be seen in the final column of Table 3, using the parameters described in this paper, a risk-based approach to inspection planning reduced costs associated with monitoring by 36.95$\%$.

\section{Discussion}\label{sec:5}

The case study showcased in Section \ref{sec:3} highlights the effectiveness of a risk-based approach to inspection planning. The significant reduction in monitoring costs can be attributed to reducing the number of unnecessary inspections while avoiding damaging the workpiece at a similar rate to a periodic inspection process. 

Deriving the equations of the expected utility of every action in the decision analysis is not always trivial. Even then, sometimes, simplifications are required if the user wishes to implement the decisions in real time, since some expected utility calculations induce large computational loads. 

While choosing a hierarchical or multilevel model to model the data in Section \ref{sec:3} provides many benefits, there are also computational considerations when working with these models. When partially pooling data in this manner, the probability space that a Monte Carlo sampler is required to explore becomes higher dimensional; this can lead to increased computational costs and restrictions in the choice of prior probability distributions (because of complex posterior geometries which can be difficult to explore). For online monitoring scenarios where decisions and actions are required instantaneously with data acquisition, increased computation times could be an issue. Additionally, the full Value of Information analysis was left out of this paper, this would  add to computation times. 

\vspace*{-12pt}
\section{Conclusion}

A Bayesian multilevel model is used to model a population of machining tools and make predictions about how the tools degrade. The equations for the expected utility of inspecting and replacing the tools were derived to form an online decision-theoretic approach to inspection planning where tools are inspected in an active manner according to the risk.

The authors believe that using risk as a query measure for active learning, rather than information measures, has a place in many engineering decision scenarios. Whilst it can be difficult to formulate the equations presented in Section \ref{sec:3} without proper understanding of the decision problem, this work shows that a risk-based approach to inspection planning can lead to a large reduction in monitoring costs while maintaining comparable or, in some cases, improved performance when compared to other methods.

\paragraph{Acknowledgments}
The authors gratefully acknowledge the support of the UK Engineering and Physical Sciences Research Council (EPSRC) via the ROSEHIPS project (Grant EP/W005816/1). CTW would like to acknowledge Dr Wayne Leahy and the team at Element Six for their expertise, resources, and materials For the purpose of open access, the authors have applied a Creative Commons Attribution (CC BY) license to any Author Accepted Manuscript version arising.

\paragraph{Funding Statement}
The authors gratefully acknowledge the support of the UK Engineering and Physical Sciences Research Council (EPSRC) via the ROSEHIPS project (Grant EP/W005816/1). For the purpose of open access, the authors have applied a Creative Commons Attribution (CC BY) license to any Author Accepted Manuscript version arising.

\paragraph{Competing Interests}
None

\paragraph{Data Availability Statement}
Requests for data will be considered on a case-by-case basis.

\paragraph{Ethical Standards}
The research meets all ethical guidelines, including adherence to the legal requirements of the study country.

\paragraph{Author Contributions}
Please provide an author contributions statement using the CRediT taxonomy roles as a guide {\verb+\url{https://www.casrai.org/credit.html}+}. Conceptualization: A.A; A.B. Methodology: A.A; A.B. Data curation: A.C. Data visualisation: A.C. Writing original draft: A.A; A.B. All authors approved the final submitted draft.

\paragraph{Supplementary Material}
State whether any supplementary material intended for publication has been provided with the submission.

\bibliographystyle{apalike}
\bibliography{main.bib}

\end{document}